\documentclass{article}
 
\usepackage{microtype}
\usepackage{graphicx}
\usepackage{subcaption}
\usepackage{booktabs}
 
\PassOptionsToPackage{hyphens}{url}
\usepackage{xurl}
\usepackage{hyperref}

\usepackage[accepted]{icml2026}
\usepackage{amsmath}
\usepackage{amssymb}
\usepackage{mathtools}
\usepackage{amsthm}

\usepackage{algorithm}
\usepackage{algorithmic}
\usepackage{xcolor}
\usepackage{array}
\usepackage{multirow}
\usepackage{enumitem}
\usepackage{pgfplots}
\pgfplotsset{compat=1.18}
\pgfplotsset{
  method/ZS/.style         ={only marks, mark=*,         mark size=2pt,  color=black},
  method/SelfRefine/.style ={only marks, mark=o,         mark size=2pt,  color=gray},
  method/SingleR/.style    ={only marks, mark=triangle*, mark size=2pt,  color=teal!70!black},
  method/TwoR/.style       ={only marks, mark=diamond*,  mark size=2.4pt,color=violet!70!black},
  method/MARS/.style       ={only marks, mark=pentagon*, mark size=2.6pt,color=red!70!black},
  method/AR/.style         ={only marks, mark=square*,   mark size=2.4pt,color=blue!70!black},
}

\icmltitlerunning{Adversarial Review: Structured Disagreement for Grounded Agentic Code Review}

\begin{document}

\twocolumn[
\icmltitle{Adversarial Review: Structured Disagreement for Grounded Agentic Code Review}

\icmlsetsymbol{equal}{*}
\begin{icmlauthorlist}
\icmlauthor{Eric S. Qiu}{equal,cornell}
\icmlauthor{Joyce Gill}{equal,stanford}
\end{icmlauthorlist}
\icmlaffiliation{cornell}{Cornell University}
\icmlaffiliation{stanford}{Stanford University}
\icmlcorrespondingauthor{Eric S. Qiu, Joyce Gill}{sq225@cornell.edu, joycegill@stanford.edu}
\icmlkeywords{LLM code review, agentic coding, multi-agent, structured disagreement, human-centered agents}
\vskip 0.3in
]

\printAffiliationsAndNotice{\icmlEqualContribution}

\begin{abstract}
Early multi-agent LLM systems often used role-separated teams, yet scaling agent count yields diminishing returns on repository-level coding tasks. Recent alternatives treat agents as passive tools (subagents), yet this removes the benefits of agent interaction entirely. We study whether a subagent paradigm can support a middle ground: minimal agentic cooperation without the overhead of large multi-agent teams. We introduce Adversarial Review (AR), a minimal cooperative code-review protocol in which a main coding agent works with a reviewer and a critic agent. The reviewer evaluates code, while the critic audits the review through structured disagreement before the main agent edits. On LiveCodeBench, AR achieves the highest pass rate among tested methods, outperforming a five-agent baseline while using only three agents. On SWE-PRBench, naive AR exposes a false-consensus failure mode, where agents converge on agreement without sufficient evidence, but a single prompt iteration that adds disagreement explicitly achieves the highest F1 among tested methods. On SWE-bench Verified, AR also shows improvements over the baselines on repository-level coding tasks. Together, AR demonstrates that cooperative code review does not require many agents or complex communication structures: it requires that disagreement be minimal, structured, and evidence-grounded.
\end{abstract}

\begin{figure*}[htb]
  \centering
  \includegraphics[width=0.9\textwidth]{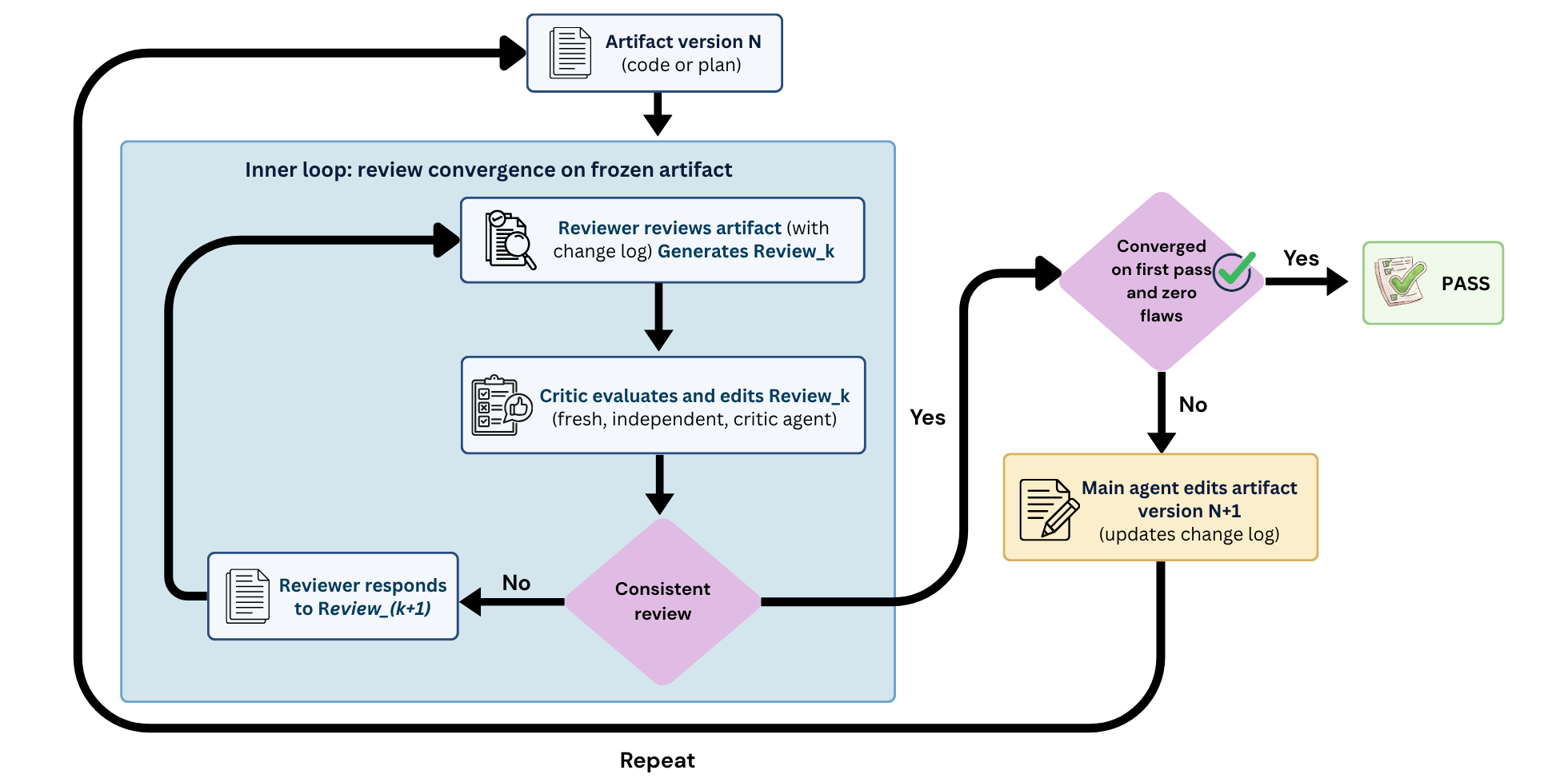}
  \caption{Workflow of Adversarial Review (AR). The main agent first produces artifact version $N$ (code or plan). The protocol then enters an inner loop in which the artifact is frozen: reviewer R generates $\mathrm{Review}_k$, critic C evaluates and may revise that review, and R responds until a consistent review is reached. If the review converges on the first pass and identifies no flaws, the artifact is accepted immediately. Otherwise, the main agent edits the artifact to produce version $N{+}1$, updates the change log, and the process repeats. The key separation is that the inner loop exchanges review text only, while artifact edits occur only in the outer loop.}
\label{fig:workflow}
\end{figure*}

\section{Introduction}
\label{sec:intro}
LLM-based coding agents are increasingly deployed to solve real software engineering tasks, from resolving GitHub issues to reviewing pull requests \citep{swebench}. As these agents grow more capable, a key design question emerges: how should multiple agents be organized to work together reliably on coding tasks? This question is especially relevant for automated code review, where an agent must not only generate correct fixes but also surface the right concerns.
 
Early multi-agent LLM systems addressed quality by adding more agents in role-separated teams, mirroring human divisions of labor by assigning agents to generate, review, debate, or aggregate outputs \citep{mapcoder, agentcoder, codecor, codesim}. However, recent work suggests that scaling the number of peer agents can yield diminishing, and sometimes negative, returns on complex repository-level coding tasks, partly because unconstrained communication introduces coordination overhead and failure modes of its own \citep{smit2023mad, scaling-agents}. In contrast, many production-grade coding agents have converged on a \emph{main agent with subagents} paradigm, where subagents are invoked as tool calls rather than as independent collaborators \citep{anthropic-agents}.

We ask whether there is a productive middle ground between these two paradigms: can a main-agent-with-subagents system support lightweight cooperation among independent agents without incurring the overhead of large role-separated teams? Our goal is to design a minimal protocol that preserves the benefits of agent interaction while constraining communication enough to remain useful in real coding tasks.

We introduce \emph{Adversarial Review} (AR), a cooperative coding protocol built step by step from naive zero-shot code generation. Each step adds one design choice motivated by a measurable failure of the previous step. We first evaluate this construction on LiveCodeBench (LCB) \citep{livecodebench2024}. We then probe the resulting protocol in two directions: review quality on SWE-PRBench \citep{swe-prbench2026} and repository-level repair on SWE-bench Verified \citep{swebench}. We use two execution modes. On LCB and SWE-PRBench, a Python orchestrator enforces strict-controlled comparisons across methods. On SWE-bench Verified, we express each method as a portable pure-text \texttt{SKILL.md} protocol followed by autonomous agents such as Claude Code.

We make three contributions. First, we show that a minimal reviewer--critic loop can improve agentic coding and program repair within the main-agent-with-subagents paradigm. Second, we identify a core failure mode of cooperative agent systems: agents can optimize for agreement rather than correctness. Third, we show that structured disagreement can reliably fix false consensus, improving both code generation and review quality across three benchmarks.

\section{Related Work}
\label{sec:related}

\paragraph{Self-refinement and external verification for code.}
Test-time refinement methods improve an initial model output by adding critique, feedback, or revision. Self-Refine uses the same model as generator, feedback provider, and refiner, keeping revision within a single model's judgment loop \citep{madaan2023self}. However, subsequent work shows that language models do not reliably self-correct reasoning errors without external feedback \citep{huang2024large}. AR builds on test-time refinement but makes two changes: it separates the reviewer and critic roles, and it makes disagreement an explicit part of the protocol rather than an optional self-check.

\paragraph{Multi-agent debate and structured disagreement.}
Multi-agent debate uses interaction among model instances by allowing agents to exchange arguments before a judge or final answer is selected to improve reasoning, factuality, or decision quality \citep{du2023mad,liang2024encouraging}. However, debate-style systems can be sensitive to prompt and protocol choices, and may fail to outperform cheaper non-debate baselines \citep{smit2023mad}. Multi-Agent Review Systems (MARS) reduces this overhead by replacing round-table debate with independent reviewers whose comments are aggregated by a meta-reviewer \citep{mars2025}. AR takes a middle position. It preserves interaction, but restricts communication to a single channel with explicit disagreement types. The goal is not open-ended debate, but evidence-grounded pushback before the main agent edits or commits.

\paragraph{Multi-agent systems for code generation and review.}
Several coding systems use role-separated agents to divide programming into specialized subtasks. MapCoder decomposes competitive-programming tasks into example recall, planning, code generation, and debugging \citep{mapcoder}. AgentCoder separates implementation, test design, and test execution \citep{agentcoder}. CodeSim uses simulation-driven planning and debugging to guide code generation \citep{codesim}. CodeCoR uses a self-reflective multi-agent framework to prune, refine, and repair generated code \citep{codecor}. For code review, CodeAgent studies communicative agents for automated review and adds a QA-Checker to keep generated review contributions aligned with the original question \citep{codeagent}. AR is intentionally smaller than these systems. Rather than adding many specialized coding roles, it asks whether a minimal reviewer--critic pair can provide reliable, evidence-grounded review within a main-agent-with-subagents workflow.

\paragraph{Agentic coding systems and portable protocols.}
Production agent systems increasingly rely on a main agent that calls tools, subagents, or reusable instruction bundles \citep{anthropic-agents}. In this setting, subagents are separate agent instances spawned for focused subtasks, while Skills are portable instruction bundles activated through a \texttt{SKILL.md} file \citep{anthropic2026subagents,anthropic2025skills}. This motivates evaluating AR in two forms: a Python orchestrator for strict-controlled comparisons, and a pure-text SKILL form that tests whether the same cooperative-review protocol can be followed by an autonomous coding agent without external enforcement. 

\paragraph{Human-agent collaboration in coding workflows.}
A growing line of work examines how coding agents should communicate progress, handle ambiguity, and incorporate user feedback \citep{codereviewer2022, swe-prbench2026}. Code review is a natural interface for human-agent collaboration: a review agent must not only find bugs but produce findings that a developer can act on, accept, or redirect. AR contributes to this agenda by making disagreement between subagents explicit, so that the review signal reaching any developer overseeing the workflow is grounded in concrete code evidence rather than unchecked speculation.

\paragraph{Benchmarks for coding and review.}
LiveCodeBench evaluates code generation using a contamination-resistant benchmark that is updated over time \citep{livecodebench2024}. SWE-PRBench evaluates AI code-review quality against pull-request feedback \citep{swe-prbench2026}. SWE-bench evaluates whether agents can resolve real GitHub issues by editing repositories and passing tests \citep{swebench}. Together, these benchmarks separate three claims that are often conflated: whether a method solves standalone programming problems, whether it helps with repository-level repair, and whether it produces useful review comments. We use this separation to evaluate AR as both a coding protocol and a cooperative-oversight protocol.

\section{Construction of AR on LiveCodeBench}
\label{sec:lcb}

This section builds the Adversarial Review protocol step by step. Each subsection adds one
design choice. Each result reports two metrics on LCB: pass-rate over
all 105 stdin-style tasks (pass/105) and pass-rate over the 57
hard-tagged subset (pass-on-hard/57). Full results are shown in \autoref{tab:lcb-headline}. The hard tier is where the methods
separate. We use Claude Sonnet 4.5 Medium Reasoning for all subsequent agent and subagent calls for all benchmarks.

\subsection{Zero-shot $\to$ Self-Refine: adding verification}
\label{sec:lcb-sr}

Zero-shot writes code with no verification step. So we add one.
Self-Refine \citep{madaan2023self} has the same agent self-critique its
own code, then revise. On LCB, the improvements are barely noticeable. Zero-shot pass-rate
is 77\%. Self-Refine pass-rate is also 77\%.  The reason is that the critic and the generator are
the same model making the same mistakes. So the critic struggles to catch what the generator missed \citep{olausson2024}. This motivates using a separate model call for the critic.

\subsection{Self-Refine $\to$ Single-reviewer: external verification}
\label{sec:lcb-singler}

Single-reviewer employs separate model call. Specifically,
main agent M writes the code, reviewer R reviews M's code, and M edits once. On LCB, pass-rate
is 77\%. Pass-on-hard is 36/57. Performance stays in the same cluster as Zero-shot. One issue may be that a single external reviewer
has high variance. It can approve everything without checking, or flag
many small nits without targeting the real issue. This motivates running multiple
independent reviewers.

\subsection{Single-reviewer $\to$ Two-reviewers: independence reduces variance}
\label{sec:lcb-tworev}

Two independent samples should reduce variance. Specifically, R$_1$ and R$_2$ review M's code independently, then M edits once
based on the union. On LCB, pass-rate is 75\%. Pass-on-hard is 34/57.
This is still within noise of Single-reviewer, Self-Refine, and Zero-shot. The four-methods cluster at the same level, and naive verification cannot push past this level. This motivates pushing for agent diversity
rather than more independent samples.

\subsection{Two-reviewers $\to$ MARS: diversity but with a limit}
\label{sec:lcb-mar}

Multi-agent debate (MAD) is the standard way to add diversity in
multi-agent LLM systems \citep{du2023mad}. However, MAD has two costs that grow with the number of
agents and rounds. The first cost is communication overhead, because
each agent must read every other agent's messages. The second cost is
that pass-rate can even drop as the system scales \citep{smit2023mad,
scaling-agents}.

MARS \citep{mars2025} reduces MAD-style communication costs by
removing message passing between reviewers. Specifically, an author
agent produces an initial solution. Three reviewers in parallel each issue a structured
output containing a binary decision (accept or reject), a
self-reported confidence score, a justification, and (on reject) an
explicit error identification. A separate meta-reviewer subagent then reads the three
reviews and the candidate solution. The meta-reviewer issues a single
consolidated output containing a final decision, a justification,
revision suggestions on reject, and a recommended answer. If the
meta-reviewer rejects, the author revises the solution using the
suggestions, and a fresh round of review runs on the revised solution.
The cycle terminates when the meta-reviewer accepts or when a fixed
cap of $K{=}2$ rounds is reached.

On LCB, MARS is
the first method whose pass-rate is clearly higher than Zero-shot,
Self-Refine, Single-reviewer, and Two-reviewers. MARS gains +5pp on
pass/105 (85\%) over the four-method cluster and +3 hard tasks (43/57) over
Single-reviewer. But MARS has two issues. First, the gain is small
relative to the token cost from using five agents per task (one author,
three reviewers, and one meta-reviewer). Second, while MARS avoids
MAD's communication overhead, all agent interactions are removed
completely. This motivates a natural question:
\textit{\textbf{can we improve coding performance through agent
interaction, yet without paying MAD's overhead, and with fewer
agents?}}

\subsection{MARS $\to$ AR: interaction with two agents}
\label{sec:lcb-ar}

A compact interactive structure that still permits meaningful
cooperation may be one reviewer plus one critic of that reviewer. We call
this \textbf{Adversarial Review (AR)}. As shown in
\autoref{fig:workflow}, the protocol operates over successive artifact
versions produced by a main agent $M$ (artifact is either code or a plan). At outer iteration $N$, $M$
produces an artifact version $N$, together with
a change log that records how the artifact evolved.

AR then enters an \emph{inner loop} in which the artifact is never edited. Instead,
an independent reviewer subagent R inspects the artifact and generates a review
$\mathrm{Review}_k$. A fresh critic subagent C then evaluates
that review and may revise or challenge it. If the review and critique
are not yet consistent, the protocol returns to R, who responds with
a revised review $\mathrm{Review}_{k+1}$. This reviewer--critic
exchange repeats until the review converges (or, for cost control,
until a cap of 5 inner rounds is reached). Thus, the inner loop is
a review procedure over a fixed artifact, not an editing
procedure over the artifact itself. This means that we can evaluate the inner loop separately as a code-review method (see \autoref{sec:prbench}).

Once a consistent review is reached, the protocol exits the inner loop
and applies an \textit{outer-loop }decision rule. If R and C converged immediately and found no flaws, AR terminates and the
artifact is accepted. This is the \emph{first-pass termination rule}.
Otherwise, the consistent review is returned to the main agent $M$,
which edits the artifact to produce version $N{+}1$, updates the change
log, and launches a new inner round of review on the updated artifact. In
this way, AR alternates between (i) an inner loop that seeks consensus
over review text of the artifact and (ii) an outer loop in which
the main agent edits the artifact only after a stable review signal has
been produced.

On LCB, AR has the highest pass-rate of all six methods we test (pass-rate is 87\%. Pass-on-hard is 43/57). In particular, AR scores +5pp
over MARS while using fewer reviewing roles: 2 (R and C) versus
MARS's 4 (three reviewers plus one meta-reviewer).

\autoref{tab:lcb-headline} summarises all six methods. Evaluation on
LCB first validates the middle ground we seek. AR shows that in the
main-agent-with-subagents paradigm, \textbf{two interacting reviewing
agents can perform better than four non-interacting reviewing agents}.

\begin{table}[t]
\centering
\small
\caption{LCB results. The leader is bold. The first four methods cluster
at the same pass-rate. MARS breaks out of the cluster. AR scores
highest with the fewest agents among the methods that broke out.\label{tab:lcb-headline}}
\begin{tabular}{l|cc|c}
\toprule
Method            & pass / 105 & pass-on-hard / 57 & \# agents \\
\midrule
Zero-shot         & 77\%       & 35/57 (61\%)      & 1 \\
Self-Refine       & 77\%       & 35/57 (61\%)      & 1 \\
Single-reviewer   & 77\%       & 36/57 (63\%)      & 2 \\
Two-reviewers     & 75\%       & 34/57 (60\%)      & 3 \\
MARS               & 82\%       & 39/57 (68\%)      & 5 \\
\textbf{AR}  & \textbf{87\%} & \textbf{43/57 (75\%)} & \textbf{3} \\
\bottomrule
\end{tabular}
\end{table}

\subsection{The compact interactive structure}
\label{sec:lcb-minimal}

Among the methods, AR has the smallest number of roles that still has interaction. AR has
three roles: M, R, and C. R and C interact. This matches the
main-plus-subagent shape used by production agent systems like Claude
Code, Cursor, and Copilot \citep{anthropic-agents}. If we remove C, the
loop becomes Single-reviewer. If we add more reviewers or more rounds,
the loop becomes MARS-style or MAD-style.

AR has the highest LCB pass-rate, but we further probe the protocol in two
directions. \autoref{sec:prbench} looks at review quality of the inner loop on
SWE-PRBench. \autoref{sec:swebench} looks at real agentic scenarios on
SWE-bench Verified.

\section{Review quality on SWE-PRBench}
\label{sec:prbench}
This section does two things. First, it shows that AR's LCB win
does not transfer naively to review-quality benchmarks. Second, it
finds the failure modes and fixes them with one prompt iteration. The
fix points to a design principle that applies to any cooperative coding agent: when agents interact to produce a shared review judgment, the protocol should force disagreement to be explicit and evidence-grounded, or the resulting review may be less reliable than a single independent reviewer.

\subsection{Experiment Setup}
\label{sec:prbench-setup}

SWE-PRBench \citep{swe-prbench2026} contains 100 real GitHub PR diffs.
The agent's job is to review the diff. A GPT-5.2 judge matches the
agent's comments against the actual human reviewer's comments. The
judge agrees with human annotators at Cohen's kappa = 0.75 (an agreement score, where 1.0 is perfect agreement and
0 is chance agreement). The metric is F1 over matched comments.

We restrict the comparison to four methods: AR, MARS, Two-reviewers,
and Single-reviewer. We exclude Zero-shot and Self-Refine, because they
do not apply to a review-only benchmark. 
Evaluating the four methods answers whether AR's
interactive R--C structure produces better reviews than methods that run reviewers in parallel.

\subsection{AR underperforms naively}
\label{sec:prbench-headline}

The naive AR protocol gives F1 = 0.457
on SWE-PRBench. This is the lowest of the four methods in the subset.
The other three subset methods all reach F1 around 0.50. So review
quality is a setting where AR's interactive R--C loop gives a lower F1
than methods without it. The rest of \autoref{sec:prbench} finds out
why and fixes it. \autoref{tab:prbench-subset} shows the leaderboard. Two case studies in
\autoref{sec:prbench-cases} surface two failure modes. \autoref{sec:prbench-v2}
describes the prompt iteration that addresses both. The iterated protocol is
\emph{AR with text constraint} and reaches F1 = 0.533, the highest in the subset.

\begin{table}[t]
\centering
\small
\caption{SWE-PRBench review-relevant subset. The leader is bold. ``AR'' and
``AR with text constraint'' share the same R+C structure; only the prompts
change. The number of agents and the way they connect did not change. The
prompt iteration is described in \autoref{sec:prbench-v2}.\label{tab:prbench-subset}}
\begin{tabular}{l|cc}
\toprule
Method                & F1  & N \\
\midrule
\textbf{AR with text constraint} & \textbf{0.533} & 100 \\
Two-reviewers         & 0.503    & 100 \\
MARS                  & 0.501    & 100 \\
Single-reviewer       & 0.495    & 100 \\
AR                    & 0.457    & 100 \\
\bottomrule
\end{tabular}
\end{table}

\subsection{Two failure modes (case studies)}
\label{sec:prbench-cases}

\paragraph{Case A — Over-decomposition.}
In AR, the reviewer R writes a review with one real bug and two to three hedged
concerns. The critic C agrees with all of R's flags and adds another
speculative bug on top. A format-review step turns each flag into a
separate comment. The judge marks 3 of 5 comments as fabricated,
because the comments are too thin or too speculative.
The mechanism is that the R--C loop adds findings without filtering
them. R hedges often. C confirms most of R's hedges. So the result is
too many thin comments, which the judge marks as fabricated. On this
task AR has F1 = 0.250.

\paragraph{Case B — C yields to R.}
In AR, the reviewer R produces an APPROVE review. The critic C raises a real concern, which we verified manually. R rebuts C with weak signals; R
uses a file-level argument yet cites no code. C gives in
and the verdict flips to AGREE. So the real bug is dropped from the
final review. Case B reveals a structural mechanism: the protocol pushes R and C to
agree, so C tends to agree. R can end the disagreement by writing a
confident-sounding rebuttal, even when R is wrong. What looks like
``structured disagreement'' becomes false agreement. On this task, AR has
F1 = 0.286, while MARS catches the same bug at F1 = 0.667.

\subsection{AR with text constraint}
\label{sec:prbench-v2}

We explicitly guide pushback constraints in text, producing AR with text constraint. Everything else stays the same. The text constraint specifies that all flags should be grounded in concrete evidence. Specifically,
C used to choose between two verdicts (\texttt{AGREE} or \texttt{DISAGREE}). Now C must
choose one of three:
\begin{itemize}[leftmargin=*,itemsep=0pt]
\item \texttt{AGREE} means C accepts R's review.
\item \texttt{DISAGREE\_EVIDENCE: <code citation>} means C cites
specific code that contradicts a flag.
\item \texttt{DISAGREE\_CONCERN: <epistemic objection>} means C raises
an objection that cannot point to contradicting code.
\end{itemize}
R's response rule depends on which kind of disagreement it sees. On
\texttt{AGREE}, R keeps every flag unchanged. On
\texttt{DISAGREE\_EVIDENCE}, R revises the flag based on the cited
code. On \texttt{DISAGREE\_CONCERN}, R must cite code in the diff that
confirms the bug (and keep the flag) or cite code that refutes the bug
(and drop the flag).

\subsection{Result}
AR with text constraint has F1 = 0.533, the highest among the four-method subset. We stop iterating to maintain a minimal protocol structure: one reviewer R, one critic C, with changes only in the prompts.

The case studies show that \emph{when two LLM agents are asked to agree on a
joint output, they tend to agree with each other. They do not always
find the truth.} So the protocol must include explicit pushback.
Without pushback, what looks like ``structured disagreement'' becomes
false agreement.

\section{Real-world coding on SWE-bench Verified}
\label{sec:swebench}

\subsection{Experimental Setup}
\label{sec:swebench-setup}

SWE-bench Verified \citep{swebench-verified} contains 500 real GitHub
issues in open-source Python repos. The agent must read the issue, navigate the repo, find the
bug, and emit a patch. The metric is pass@1, measured through the
official Docker harness.

The execution mode here is different from \autoref{sec:lcb} and
\autoref{sec:prbench}. We use Claude Code with full tool access. Claude
Code reads the \texttt{adversarial-review} skill, which is simply a markdown doc (SKILL.md) that records the AR protocol in pure text (rather than Python orchestration logic). 

Evaluation of the skills form of AR has \textit{desirable characteristics}. First, the skills form, by design, let a main LLM agent drive the protocol itself. Specifically, the agent can decide when to invoke R and C as Task
subagents, and it also decides when to edit. While our evaluation harness of AR for LCB and SWE-PRBench intentionally used Python orchestrating logic for strict execution comparisons, this leans more towards an \textit{agentic workflow}, where the structure is rigid. On the other hand, a more general notion of LLM agent can decide when to invoke and how to structure its own workflow. Coding agents, which often need flexibility and innovative decision-making to solve complex coding tasks, fall under the latter notion of agents. Hence, the skills form of AR provides a natural way to evaluate the protocol on flexible coding agents. 

We evaluate only three methods on SWE-bench Verified because full benchmark runs estimate over 300 hours for each method. We choose Zero-shot as the naive Claude Code baseline, MARS as the strongest tested multi-agent baseline, and AR as our proposed method. The MARS body used here is the same protocol as in \autoref{sec:lcb-mar} and \autoref{sec:prbench}, expressed as an analogous SKILL-style document so the agent can follow it autonomously; see \autoref{app:prompts-swebench-mars-body}.

\subsection{Result}
\label{sec:swebench-headline}

As shown in \autoref{tab:swebench}, AR reaches 75.2\% pass-rate on N=500.
Zero-shot on the same Claude Code scaffold reaches 71.6\%.
MARS reaches 72.6\%. Though notably, AR uses about 4.5$\times$ the tokens of
Zero-shot (see \autoref{sec:cost}), trading off token-efficiency for performance
gains.

\begin{table}[h]
\centering
\small
\caption{SWE-bench Verified pass-rate of three methods: Zero-shot, MARS, and AR.}
\label{tab:swebench}
\begin{tabular}{l|cc}
\toprule
Method            & pass-rate (\%) & N \\
\midrule
\textbf{AR}       & \textbf{75.2\%} & 500 \\
Zero-shot         & 71.6\%  & 500 \\
MARS              & 72.6\%  & 500 \\
\bottomrule
\end{tabular}
\end{table}

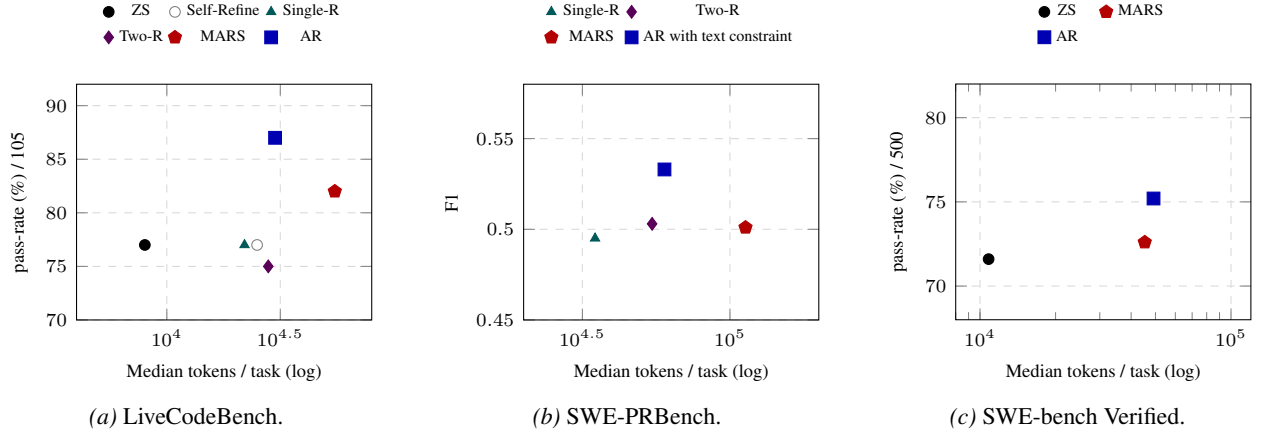
\begin{figure*}[h]
\centering
\begin{subfigure}[t]{0.32\textwidth}
\vspace{0pt}\centering
\begin{tikzpicture}
\begin{axis}[
  width=\textwidth, height=4.7cm,
  xlabel={Median tokens / task (log)},
  ylabel={pass-rate (\%) / 105},
  xmode=log, log basis x=10,
  xmin=4000, xmax=80000,
  ymin=70, ymax=92,
  grid=major, grid style={dashed,gray!30},
  clip=false,
  legend columns=3,
  legend style={font=\tiny, at={(0.5,1.12)}, anchor=south, draw=none},
  tick label style={font=\scriptsize},
  label style={font=\scriptsize},
]
\addplot[method/ZS] coordinates {(8000, 77)};
\addlegendentry{ZS}
\addplot[method/SelfRefine] coordinates {(25000, 77)};
\addlegendentry{Self-Refine}
\addplot[method/SingleR] coordinates {(22000, 77)};
\addlegendentry{Single-R}
\addplot[method/TwoR] coordinates {(28000, 75)};
\addlegendentry{Two-R}
\addplot[method/MARS] coordinates {(55000, 82)};
\addlegendentry{MARS}
\addplot[method/AR] coordinates {(30000, 87)};
\addlegendentry{AR}
\end{axis}
\end{tikzpicture}
\caption{LiveCodeBench.\label{fig:pareto-lcb}}
\end{subfigure}\hfill
\begin{subfigure}[t]{0.32\textwidth}
\vspace{0pt}\centering
\begin{tikzpicture}
\begin{axis}[
  width=\textwidth, height=4.7cm,
  xlabel={Median tokens / task (log)},
  ylabel={F1},
  xmode=log, log basis x=10,
  xmin=20000, xmax=200000,
  ymin=0.45, ymax=0.58,
  grid=major, grid style={dashed,gray!30},
  clip=false,
  legend columns=2,
  legend style={font=\tiny, at={(0.5,1.12)}, anchor=south, draw=none},
  tick label style={font=\scriptsize},
  label style={font=\scriptsize},
]
\addplot[method/SingleR] coordinates {(34925, 0.495)};
\addlegendentry{Single-R}
\addplot[method/TwoR] coordinates {(54492, 0.503)};
\addlegendentry{Two-R}
\addplot[method/MARS] coordinates {(112902, 0.501)};
\addlegendentry{MARS}
\addplot[method/AR] coordinates {(60000, 0.533)};
\addlegendentry{AR with text constraint}
\end{axis}
\end{tikzpicture}
\caption{SWE-PRBench.\label{fig:pareto-prbench}}
\end{subfigure}\hfill
\begin{subfigure}[t]{0.32\textwidth}
\vspace{0pt}\centering
\begin{tikzpicture}
\begin{axis}[
  width=\textwidth, height=4.7cm,
  xlabel={Median tokens / task (log)},
  ylabel={ pass-rate (\%) / 500},
  xmode=log, log basis x=10,
  xmin=8000, xmax=120000,
  ymin=68, ymax=82,
  grid=major, grid style={dashed,gray!30},
  clip=false,
  legend columns=2,
  legend style={font=\tiny, at={(0.5,1.12)}, anchor=south, draw=none},
  tick label style={font=\scriptsize},
  label style={font=\scriptsize},
]
\addplot[method/ZS] coordinates {(10820, 71.6)};
\addlegendentry{ZS}
\addplot[method/MARS] coordinates {(45320, 72.6)};
\addlegendentry{MARS}
\addplot[method/AR] coordinates {(49126, 75.2)};
\addlegendentry{AR}
\addlegendimage{empty legend}
\addlegendentry{}
\end{axis}
\end{tikzpicture}
\caption{SWE-bench Verified.\label{fig:pareto-swebench}}
\end{subfigure}
\caption{Per-task median tokens vs.\ quality on each benchmark. The three y-axes
measure different things across panels (binary pass-rate vs.\ LLM-judge F1).
The within-panel claim is the within-panel trade-off, not a cross-panel
comparison. LCB token medians are estimated from per-method call counts;
SWE-PRBench and SWE-bench Verified token medians are computed from per-task
logs.\label{fig:pareto}}
\end{figure*}

\subsection{Case studies}
\label{sec:swebench-cases}

We use two case studies to further analyze AR's results on SWE-bench Verified. The
first case shows when AR helps: the reviewer--critic loop pushes the
agent toward the root cause. The second case shows when AR hurts: the
same loop amplifies a speculative concern and expands the patch beyond
the issue scope. Both cases come from AR's skills-form evaluation.

\paragraph{Case 1 — Structured disagreement helps find the root cause.}
\textit{Task: \texttt{matplotlib\_\_matplotlib-20826}.}
The issue occurs after \texttt{Axis.clear()} resets the per-tick keyword
argument dictionaries. As a result, when a user calls
\texttt{ax.clear()} on shared or subplot axes, tick-visibility settings
such as labels hidden by \texttt{label\_outer()} are lost.

Zero-shot and MARS both patch the caller:
\texttt{lib/matplotlib/axes/\_base.py}. Their patches re-apply tick
visibility settings after the reset has already destroyed them. This
fixes the symptom, but not the underlying cause. Zero-shot produces a
45-line patch, and MARS produces a 50-line patch. Both fail the hidden
tests.

In comparison, AR instead patches the callee:
\texttt{lib/matplotlib/axis.py}. Rather than re-applying visibility
settings after the reset, AR removes the over-broad reset and preserves
the relevant tick keyword state directly:

\begin{quote}\small\ttfamily
- self.\_reset\_major\_tick\_kw() \\
- self.\_reset\_minor\_tick\_kw() \\
+ self.\_major\_tick\_kw['gridOn'] = ... \\
+ self.\_minor\_tick\_kw['gridOn'] = ...
\end{quote}

This difference matters. Zero-shot and MARS repair the visible symptom
at the public-API layer. While AR's reviewer subagent R makes the same mistake in the initial round of reviews, the critic subagent C catches the mistake right away, and the agents converge to agreement immediately. As a result, AR actually repairs the state-destroying operation at
the implementation layer. AR's patch is shorter, about 20 lines, and is
the only patch that passes the hidden tests. We interpret this as a case
where the reviewer--critic loop helps the main agent inspect one frame
deeper in the call chain, moving from symptom repair to root-cause
repair.

\paragraph{Case 2 — Structured disagreement amplifies scope creep.}
\textit{Task: \texttt{astropy\_\_astropy-14182}.}
The issue asks the agent to add a \texttt{header\_rows} argument to the
RST writer. Zero-shot solves the task with a minimal 24-line patch that
threads the parameter through \texttt{\_\_init\_\_} and \texttt{write}.
The patch stays within the issue scope and passes the hidden tests.

AR produces a larger 32-line patch. It adds \texttt{header\_rows}, but
also removes a stable class variable,
\texttt{start\_line=3} in \texttt{SimpleRSTData}, and adds a new
\texttt{read()} method. These extra changes were not required by the
issue. The source of the over-edit is visible in the trace: during
planning, the reviewer asks, ``what if the user calls
\texttt{read()} separately?'' The critic does not reject this as out of
scope. The main agent then implements the speculative change, and the
patch fails the hidden tests.

This case shows the main failure mode of AR on SWE-bench Verified. The
reviewer--critic loop can improve root-cause analysis, but it can also
turn plausible concerns into unnecessary edits. In other words, AR helps
when disagreement redirects the agent toward the true bug, and hurts
when disagreement expands the repair beyond the issue. This suggests
that future versions of AR could make scope discipline as explicit as
the evidence-grounded disagreement we encoded in AR with text constraint (\autoref{sec:prbench-v2}).

\section{Cost-quality position across benchmarks}
\label{sec:cost}

Though AR trades token cost for performance, as we see in \autoref{sec:swebench-headline}, nevertheless AR is on the cost-quality Pareto frontier of all three benchmarks
we evaluate. The Pareto frontier is the set of methods where no other
method is both cheaper and better. \autoref{fig:pareto} plots per-task
median tokens against performance on each benchmark.

The takeaway is that AR uses more tokens than Zero-shot on all
three benchmarks. The extra tokens give AR Pareto-frontier
performance on every panel. So no method we tested is both
cheaper and better than AR.

\section{Discussion and limitations}
\label{sec:limitations}

\paragraph{AR works because it makes cooperation narrow.}
The main result of this paper is not that adding more agents improves
coding. In fact, our constructive progression shows the opposite:
adding independent reviewers alone does not reliably improve LCB, and
MARS requires five agents (one author, three reviewers, and one meta-reviewer) to obtain a smaller gain than AR. The useful
part of AR is more specific. AR restricts cooperation to one narrow
interface: a reviewer writes a review, and a critic audits that review.
The artifact is frozen during this exchange, and edits occur only after the
review stabilizes. This separation matters because it prevents the
agents from jointly rewriting the solution while they are still
disagreeing about what is wrong. In this sense, while AR is a small oversight mechanism inserted into a main-agent workflow, AR behaviorally resolves as a collaborative agent team with distinct role separation.

\paragraph{The results support a middle ground between tools and teams.}
Production coding agents often follow a main-agent-plus-tools pattern,
where subagents are invoked as tools. Earlier multi-agent coding systems
often follow a team pattern, where many role-separated agents generate and aggregate outputs. AR sits between these two designs. The
main agent remains responsible for editing the artifact,
but the reviewer and critic are not merely passive tools. They interact
with each other before the main agent acts. This design gives AR some
benefits of role-separated cooperation without requiring a large
round-table debate like MAD \citep{du2023mad, smit2023mad, mars2025}. The SWE-bench Verified result is important for this
reason: AR still improves performance when expressed only as a
\texttt{SKILL.md} protocol, without a Python orchestrator forcing every
step. This suggests that lightweight cooperative protocols can be
portable across agentic systems when the roles and decision rules are
simple enough to express in text.

\paragraph{False consensus is a reliability failure, not just a performance failure}
The SWE-PRBench results show why cooperative AI systems need to be
audited, not merely benchmarked. AR performs best on LCB but worst
among the review-centered methods on SWE-PRBench. The case studies make
the failure visible. In one case, the reviewer proposes too many weak
findings and the critic confirms them. In another case, the critic
raises a real concern, but then yields to a confident rebuttal from the
reviewer. Both failures have the same structure: the agents reach
agreement, but the agreement is not supported by enough evidence. We
call this failure mode \emph{false consensus}. False consensus is
especially concerning because it can look like independent validation:
two agents appear to agree, but the agreement may only reflect
conversational pressure to converge. This failure mode reveals when good performance on one benchmark (code generation) does not transfer to good performance on a related task (code review). Additionally, the failure mode is \emph{structural}---it arises from how agents interact, not from individual agent quality. AR addresses this by explicitly showing disagreement, which forces evidence-grounded pushback before any flag can be settled.

\paragraph{Evidence-grounded disagreement is a key design.}
AR with text constraint improves SWE-PRBench performance by adding textual constraints
rather than more agents. The critic must distinguish agreement,
evidence-backed disagreement, and concern-based disagreement. The
reviewer must then respond with code evidence instead of a plausible
rebuttal, which turns disagreement into an auditable object. This is the main trustworthy-AI lesson of AR:
multi-agent oversight should not be judged only by whether agents
converge. It should be judged by whether the path to convergence
preserves dissent, evidence, and accountability. Our textual constraint forces an agent to discretize its disagreement into exactly one of three types, which makes agent disagreement clear-cut, explicit, and transparent. However, textual constraint may not be the best nor the final solution to enforcing semantic constraints in LLM agents. While the instruction-following capability of LLMs continue to grow along its general ability, current LLMs are still prone to forgetting or ignoring instructions, especially as the context window approaches full and past full. Further work on methods to guide LLMs is an orthogonal research direction to our work, yet nevertheless important to creating trustworthy AI.

\paragraph{Cost trade-off.}
AR is not a free improvement. It uses more tokens than Zero-shot on all
three benchmarks. The cost-quality plots show that AR lies on the
Pareto frontier among the methods we tested, but this does not mean AR
should always be used. For easy tasks, the review loop may be wasted
computation. For tasks where the issue scope is narrow, extra review may
increase the chance of over-editing. AR is most justified when the repository is complex, or when
root-cause localization is more important than minimizing token cost.
This suggests a natural future direction: an adaptive version of AR that
invokes the reviewer--critic loop only when the main agent is uncertain
or when the patch touches high-risk code.

\paragraph{Our claims about agent cooperation are empirical, not formal.}
We do not claim that
reviewer--critic loops are universally better than independent
reviewers, debate, or single-agent refinement. Our conclusions are based
on three empirical probes: constructive progression on LCB, inner-loop
review quality on SWE-PRBench, and skills-form repository repair on
SWE-bench Verified. These settings are complementary, but they do not
cover all coding tasks or all agent scaffolds. In particular, the
Python-orchestrated experiments isolate the protocol under controlled execution, while the SWE-bench Verified experiment tests a looser
skills-form instruction. These two modes answer different questions.
The first mode asks whether the mechanism works under strict workflow.
The second asks whether the mechanism survives when delegated to an
autonomous  agent.

\paragraph{The evaluation is limited by benchmarks and judge design.}
LCB measures competitive-programming style correctness, SWE-PRBench
measures similarity to human pull-request feedback, and SWE-bench
Verified measures repository-level repair through hidden tests. Each
benchmark captures only one view of quality. In particular,
SWE-PRBench depends on an LLM judge that matches generated comments to
human review comments. This makes it useful for comparing review
methods, but it may penalize valid comments that differ from the human
review or reward comments that match surface form without improving the
patch. Similarly, SWE-bench Verified measures whether the final patch
passes tests, but it does not fully measure maintainability, minimality,
or long-term code quality. The case studies partly address this gap by
examining mechanism, but larger and more comprehensive evaluations would be needed to
validate AR as a deployed review assistant.

\paragraph{Future work should study adaptive and auditable cooperation.}
The next step for cooperative coding agents is better control over \emph{when} the reviewer--critic loop is invoked and \emph{how} it interacts with the main agent's planning. For AR specifically, variants could add a scope-checking critic, a confidence gate that skips review on easy tasks, or training signal derived from disagreement traces to improve agent alignment for code. More broadly, our results suggest that review protocols should be designed around explicit failure modes rather than around agent count or interaction rounds.

\section{Conclusion}
\label{sec:conclusion}

We construct AR step by step on LCB. The progression goes Zero-shot,
Self-Refine, Single-reviewer, Two-reviewers, MARS, AR. AR achieves the highest measured score among the methods we test on all three benchmarks, while remaining a compact interactive structure in the main-agent-with-subagents paradigm. Our key finding is that rather than naively scaling agents count, designing structured, explicit, evidence-grounded disagreement may more effectively improve both code generation and code review quality. 

\section*{Impact Statement}
This paper presents work whose goal is to advance the field of Machine
Learning. There are many potential societal consequences of our work, none
which we feel must be specifically highlighted here.
\begingroup
\sloppy
\setlength{\emergencystretch}{2em}
\bibliographystyle{icml2026}
\bibliography{paper}
\endgroup

\appendix
\onecolumn

\section{Prompt templates}
\label{app:prompts}

This appendix dumps every prompt used by every method, grouped by
benchmark. Python-style format-string placeholders such as
\texttt{\{specification\}}, \texttt{\{entry\_point\}},
\texttt{\{frozen\_ask\}}, \texttt{\{diff\}},
\texttt{\{problem\_statement\}}, \texttt{\{reviewer\_latest\}}, and
\texttt{\{prior\_exchange\}} are substituted at run time by the
orchestrator. The prompts will be released alongside the
implementation upon acceptance.

\subsection{LiveCodeBench (LCB) -- end-to-end code generation}
\label{app:prompts-lcb}

\subsubsection{All methods: main-agent first-version write (LCB)}
\label{app:prompts-lcb-m-write}
\label{app:prompts-lcb-write}

Used by every method to produce the initial code.

\begin{footnotesize}
\begin{verbatim}
You are writing a Python function to satisfy a specification.

## Specification
{specification}

## Entry point (MANDATORY function name)
Your function MUST be named exactly `{entry_point}`. The hidden test harness
calls the function by that literal name -- if you rename it, every test fails
with NameError regardless of semantic correctness.

## Previous consistent review (if outer iteration > 1)
{previous_review}

## Your task
Write the complete Python code. Emit ONLY the code -- no explanation, no prose,
no markdown fences (do NOT wrap in ```python ... ``` or ``` ... ```). The code
will be run verbatim against hidden tests.

Include all necessary imports at the top. If the specification references
helper utilities without defining them, implement them inline.
\end{verbatim}
\end{footnotesize}

\subsubsection{All methods: main-agent edit-from-review (LCB)}
\label{app:prompts-lcb-m-edit}

Used by every review-having method to revise code in response to a review.

\begin{footnotesize}
\begin{verbatim}
You are revising a Python function based on a code review.

## Specification
{specification}

## Entry point (MANDATORY function name)
Your function MUST be named exactly `{entry_point}`. The hidden test harness
calls the function by that literal name.

## Previous version of your code
```python
{previous_artifact}
```

## Code review (consistent review from reviewer-critic loop)
{consistent_review}

## Your task
Rewrite the COMPLETE updated Python code addressing the review. Emit ONLY the
code -- no explanation, no prose, no markdown fences.
\end{verbatim}
\end{footnotesize}

\subsubsection{Zero-shot (LCB)}

Uses \autoref{app:prompts-lcb-write} only. No reviewer or critic prompts.

\subsubsection{Self-Refine: self-critique (LCB)}

\begin{footnotesize}
\begin{verbatim}
You wrote this code:
```python
{code}
```
Critique your own code for correctness only. Do not rewrite. List specific
bugs, edge-case misses, or logical errors. End with a verdict line on its own:
  NO_FURTHER_ISSUES
or
  NEEDS_REVISION: <one-line summary>
\end{verbatim}
\end{footnotesize}

\subsubsection{Self-Refine: revise-from-critique (LCB)}

\begin{footnotesize}
\begin{verbatim}
Task specification:
{specification}

You wrote this code:
```python
{code}
```

You then critiqued your own code as follows:
{critique}

Now rewrite the COMPLETE code addressing every flagged issue. Emit ONLY
the code -- no prose, no markdown fences.
\end{verbatim}
\end{footnotesize}

\subsubsection{Single-reviewer, Two-reviewers, and AR: reviewer prompt (LCB)}
\label{app:prompts-reviewer}

\begin{footnotesize}
\begin{verbatim}
You are a senior software engineer reviewing a code artifact.

## Frozen original ask (verbatim -- do not paraphrase)
{frozen_ask}

## Artifact to review
```python
{artifact_code}
```

## Previous inner-loop exchange (if any)
{prior_exchange}

## Your task
Review the code for correctness. Flag bugs, edge cases, and logic errors ONLY.
Do not comment on style. Be specific -- cite line numbers or code snippets.

## Output format
End your review with a verdict line on its own, one of:

  APPROVE

or

  NEEDS_CHANGES: <one-line summary of top issue>

Any other final line is invalid.
\end{verbatim}
\end{footnotesize}

\subsubsection{AR: critic prompt (LCB)}

\begin{footnotesize}
\begin{verbatim}
You are critiquing a fellow engineer's code review.

## Frozen original ask (verbatim)
{frozen_ask}

## Artifact being reviewed
```python
{artifact_code}
```

## Reviewer's latest review
{reviewer_latest}

## Prior inner-loop exchange (if any)
{prior_exchange}

## Your task (TWO dimensions)
1. Is every issue the reviewer flagged actually a real bug? Flag spurious ones.
2. Did the reviewer miss any real bugs? Name them.

## Output format
End your response with a verdict line on its own, one of:

  AGREE

(if you agree with the reviewer and name no additional missed bugs)

or

  DISAGREE: <one-line summary of what you disagree with or what's missed>

Any other final line is invalid.
\end{verbatim}
\end{footnotesize}

\subsubsection{MARS: author prompt (LCB)}
\label{app:prompts-mars-author}

MARS reuses the existing main-agent prompts: the first-version write
of \autoref{app:prompts-lcb-m-write} on round~1, and the edit-from-review
of \autoref{app:prompts-lcb-m-edit} on subsequent rounds with the
meta-reviewer's Suggestions field filled into the
\texttt{\{consistent\_review\}} slot.

\subsubsection{MARS: reviewer prompt (LCB)}
\label{app:prompts-mars-reviewer-lcb}

We design a separate reviewer prompt for MARS to stay faithful to the 
structure of the originalMARS paper. The artifact under
review is a generated Python program. Each round dispatches three
reviewer subagents in parallel; they do not see each other's
outputs. The four-field output schema (Decision, Confidence,
Justification, Errors) is faithful to \citet{mars2025} and is
identical to the SWE-PRBench and SWE-bench Verified MARS reviewer
prompts of \autoref{app:prompts-mars-reviewer-prbench} and
\autoref{app:prompts-swebench-mars-body}; the only differences across
benchmarks are the artifact framing and the slots filled in.

\begin{footnotesize}
\begin{verbatim}
You are a reviewer in the MARS protocol. You evaluate a candidate
Python program INDEPENDENTLY of any other reviewer.

## Frozen original ask (verbatim -- do not paraphrase)
{frozen_ask}

## Candidate program
```python
{artifact_code}
```

## Your task
1. Read the original ask and form an independent understanding of
   what a correct program would do, BEFORE reading the candidate.
2. Analyze the candidate against your understanding.
3. Check: does the candidate satisfy the original ask? Are there
   edge cases missed? Are there logic errors? Is it scoped to the
   ask (no unrelated additions)?

## Output format (strict -- exactly these four labelled fields)

Decision: [accept | reject]
Confidence: [1-5]   (self-reported; 5 = certain)
Justification: [evidence-based reasoning; cite line numbers or code
                snippets when applicable]
Errors: [if reject: explicit list of errors in the candidate.
         If accept: leave blank.]
\end{verbatim}
\end{footnotesize}

\subsubsection{MARS: meta-reviewer prompt (LCB)}
\label{app:prompts-mars-meta-lcb}

The meta-reviewer is dispatched once per round after the three
reviewers complete. It reads the candidate program, the three
independent reviews, and issues a single consolidated decision plus,
on reject, concrete revision suggestions for the author. The
four-field output schema (Decision, Justification, Suggestions,
Recommended-answer) is faithful to \citet{mars2025} and matches the
SWE-PRBench and SWE-bench Verified meta-reviewer prompts of
\autoref{app:prompts-mars-meta-prbench} and
\autoref{app:prompts-swebench-mars-body}.

\begin{footnotesize}
\begin{verbatim}
You are the meta-reviewer in the MARS protocol. You received a
candidate Python program and three independent reviews of it. Your
job is to issue a single consolidated decision plus, on reject,
concrete revision suggestions for the author.

## Frozen original ask (verbatim)
{frozen_ask}

## Candidate program
```python
{artifact_code}
```

## Independent reviews
--- Reviewer 1 ---
{review_1}

--- Reviewer 2 ---
{review_2}

--- Reviewer 3 ---
{review_3}

## Critical instructions
1. Do NOT count votes. Weigh the reasoning and evidence each
   reviewer provides.
2. Confidence scores are signals, not deciders. A reviewer can be
   confidently wrong.
3. If reviewers disagree, prefer the position grounded in concrete
   evidence about the candidate program.

## Output format (strict -- exactly these four labelled fields)

Decision: [accept | reject]
Justification: [your reasoning, weighing the three reviews]
Suggestions: [if reject: concrete, actionable revision instructions
              for the author. If accept: leave blank.]
Recommended-answer: [a short summary of the correct program as you
                     understand it]
\end{verbatim}
\end{footnotesize}

\subsection{SWE-PRBench -- review-only inner loop}
\label{app:prompts-prbench}

SWE-PRBench is review-only: a real PR diff is the artifact, and the
reviewer/critic produce a review of that diff. There is no main-agent
edit step. Two prompt variants of AR are reported: AR (baseline; F1 = 0.457)
and AR with text constraint (F1 = 0.533; \autoref{sec:prbench-v2}).

\textbf{Reviewer prompt sharing on SWE-PRBench.}
Single-reviewer, Two-reviewers, and AR all use the reviewer prompt
below (\autoref{app:prompts-prbench-v1-r}); they differ in how the
prompt is invoked (one call for Single-reviewer; two independent
calls for Two-reviewers; one call as part of an R--C inner loop for
AR). \textbf{MARS does NOT use this prompt.} MARS instead uses
its own SWE-PRBench-flavored reviewer prompt
(\autoref{app:prompts-mars-reviewer-prbench}) followed by its own
SWE-PRBench-flavored meta-reviewer prompt
(\autoref{app:prompts-mars-meta-prbench}). The protocol runs three
independent reviewers and one meta-reviewer in one round ($K{=}1$);
because SWE-PRBench is review-only and has no author revision step,
the meta-reviewer's consolidated output is the review submitted to
the judge. AR with text constraint swaps in the text-constrained reviewer
(\autoref{app:prompts-prbench-v2-r}) and the text-constrained critic
(\autoref{app:prompts-prbench-v2-c}).

\subsubsection{Single-reviewer, Two-reviewers, and AR: reviewer prompt (SWE-PRBench)}
\label{app:prompts-prbench-v1-r}

\begin{footnotesize}
\begin{verbatim}
You are a senior software engineer performing a pull-request code review.

## Frozen original ask (verbatim)
{frozen_ask}

## Problem statement (what the PR aims to fix)
{problem_statement}

## Code diff / PR patch being reviewed
```diff
{diff}
```

## Previous inner-loop exchange (if any)
{prior_exchange}

## Your task
Identify bugs, logic errors, and missing edge-case handling IN THIS DIFF.
If the diff looks correct, say so. Be specific -- cite file:line or code
snippets when flagging an issue.

## Output format
End your review with a verdict line, one of:

  APPROVE

or

  NEEDS_CHANGES: <one-line summary of top issue>
\end{verbatim}
\end{footnotesize}

\subsubsection{AR: critic prompt (SWE-PRBench)}
\label{app:prompts-prbench-v1-c}

\begin{footnotesize}
\begin{verbatim}
You are critiquing a fellow engineer's code review of a pull request.

## Frozen original ask (verbatim)
{frozen_ask}

## Problem statement
{problem_statement}

## Code diff being reviewed
```diff
{diff}
```

## Reviewer's latest review
{reviewer_latest}

## Prior inner-loop exchange (if any)
{prior_exchange}

## Your task (TWO dimensions)
1. Is every issue the reviewer flagged actually a real bug? Flag spurious ones.
2. Did the reviewer miss any real bugs? Name them.

## Output format
End your response with:

  AGREE

or

  DISAGREE: <summary>
\end{verbatim}
\end{footnotesize}

\subsubsection{AR with text constraint: reviewer prompt (SWE-PRBench)}
\label{app:prompts-prbench-v2-r}

The only diff from baseline AR is the new ``\#\# When responding to the critic''
block, which encodes the response rule per critic-verdict type. The
rest of the prompt is identical to baseline AR.

\begin{footnotesize}
\begin{verbatim}
You are a senior software engineer performing a pull-request code review.

## Frozen original ask (verbatim)
{frozen_ask}

## Problem statement (what the PR aims to fix)
{problem_statement}

## Code diff / PR patch being reviewed
```diff
{diff}
```

## Previous inner-loop exchange (if any)
{prior_exchange}

## Your task
Identify bugs, logic errors, and missing edge-case handling IN THIS DIFF.
If the diff looks correct, say so. Be specific -- cite file:line or code
snippets when flagging an issue.

## When responding to the critic
The critic emits one of three verdict types. Respond differently to each:

  AGREE
    The critic accepts your review. Keep ALL your flags unchanged.
    Do NOT consolidate, drop, or rephrase legitimate findings just because
    you are about to converge -- preserve them as-is.

  DISAGREE_EVIDENCE: <code citation>
    The critic cites code that contradicts a flag. You were probably wrong
    on the specifics. Revise the flag based on the evidence, or drop it.

  DISAGREE_CONCERN: <epistemic objection>
    The critic doubts a flag but cannot cite contradicting code. Your job
    is EITHER to cite specific code in the diff that confirms the bug
    (keep the flag, firm it up) OR to cite specific code that refutes it
    and drop the flag. Do NOT capitulate to APPROVE merely because the
    critic called the concern "speculative" -- firm it up or rebut it
    with code.

If the critic added a missed bug (under any verdict type), incorporate it.

## Output format
End your review with a verdict line, one of:

  APPROVE

or

  NEEDS_CHANGES: <one-line summary of top issue>
\end{verbatim}
\end{footnotesize}

\subsubsection{AR with text constraint: critic prompt (SWE-PRBench)}
\label{app:prompts-prbench-v2-c}

The only diff from baseline AR is the new ``\#\# Output format -- three verdict types'' block, which allows for three instead of two choices of verdicts. The rest of the prompt is identical to baseline AR.

\begin{footnotesize}
\begin{verbatim}
You are critiquing a fellow engineer's code review of a pull request.

## Frozen original ask (verbatim)
{frozen_ask}

## Problem statement
{problem_statement}

## Code diff being reviewed
```diff
{diff}
```

## Reviewer's latest review
{reviewer_latest}

## Prior inner-loop exchange (if any)
{prior_exchange}

## Your task (TWO dimensions)
1. Is every issue the reviewer flagged actually a real bug? Flag spurious ones.
2. Did the reviewer miss any real bugs? Name them.

## Output format -- three verdict types
End your response with EXACTLY ONE of these verdict lines:

  AGREE
    Use when every flag is real and you have nothing to add.

  DISAGREE_EVIDENCE: <cite file:line or code snippet that contradicts the flag>
    Use when you can ground your objection in code visible in the diff.

  DISAGREE_CONCERN: <flag is plausible but not yet substantiated>
    Use when your objection is epistemic (reviewer hedged, you have a gut
    reaction, external attribution argument) but you cannot point to code
    that refutes the flag. The reviewer's job on the next round is then to
    firm up the flag with evidence or drop it.

Use DISAGREE_EVIDENCE when you have code grounding; DISAGREE_CONCERN when
you only have epistemic doubt. Do NOT use DISAGREE_CONCERN as a dismissal
mechanism -- use it to request evidence, not to suppress findings.
\end{verbatim}
\end{footnotesize}

\subsubsection{MARS: reviewer prompt (SWE-PRBench)}
\label{app:prompts-mars-reviewer-prbench}

This is the SWE-PRBench-flavored reviewer prompt for MARS. The
artifact under review is a real PR diff. Each round dispatches three
reviewer subagents in parallel; they do not see each other's
outputs. The four-field output schema is identical to the LCB version
(\autoref{app:prompts-mars-reviewer-lcb}); only the artifact framing
and slots differ.

\begin{footnotesize}
\begin{verbatim}
You are a reviewer in the MARS protocol, performing a pull-request
code review INDEPENDENTLY of any other reviewer.

## Frozen original ask (verbatim)
{frozen_ask}

## Problem statement (what the PR aims to fix)
{problem_statement}

## Code diff / PR patch under review
```diff
{diff}
```

## Your task
1. Read the problem statement and form an independent understanding
   of what a correct fix would do, BEFORE reading the diff.
2. Analyze the diff against your understanding.
3. Check: does the diff address the problem? Are there edge cases
   missed? Does it introduce regressions? Is it scoped to the
   problem (no unrelated refactors)?

## Output format (strict -- exactly these four labelled fields)

Decision: [accept | reject]
Confidence: [1-5]   (self-reported; 5 = certain)
Justification: [evidence-based reasoning; cite file:line or code
                snippets when applicable]
Errors: [if reject: explicit list of errors in the diff.
         If accept: leave blank.]
\end{verbatim}
\end{footnotesize}

\subsubsection{MARS: meta-reviewer prompt (SWE-PRBench)}
\label{app:prompts-mars-meta-prbench}

The meta-reviewer is dispatched once after the three reviewers
complete. It reads the PR diff, the three independent reviews, and
issues a single consolidated review. Because SWE-PRBench is
review-only and has no author revision step, the meta-reviewer's
output (specifically the Justification and Errors-style content) is
the review submitted to the judge; the Suggestions and
Recommended-answer fields are still produced for protocol fidelity
but not scored. The four-field output schema matches the LCB version
(\autoref{app:prompts-mars-meta-lcb}); only the artifact framing and
slots differ.

\begin{footnotesize}
\begin{verbatim}
You are the meta-reviewer in the MARS protocol. You received a PR
diff and three independent reviews of it. Your job is to issue a
single consolidated decision plus, on reject, concrete revision
suggestions.

## Frozen original ask (verbatim)
{frozen_ask}

## Problem statement
{problem_statement}

## Code diff / PR patch under review
```diff
{diff}
```

## Independent reviews
--- Reviewer 1 ---
{review_1}

--- Reviewer 2 ---
{review_2}

--- Reviewer 3 ---
{review_3}

## Critical instructions
1. Do NOT count votes. Weigh the reasoning and evidence each
   reviewer provides.
2. Confidence scores are signals, not deciders. A reviewer can be
   confidently wrong.
3. If reviewers disagree, prefer the position grounded in concrete
   evidence about the diff.

## Output format (strict -- exactly these four labelled fields)

Decision: [accept | reject]
Justification: [your reasoning, weighing the three reviews]
Suggestions: [if reject: concrete, actionable revision instructions.
              If accept: leave blank.]
Recommended-answer: [a short summary of the correct fix as you
                     understand it]
\end{verbatim}
\end{footnotesize}

\subsection{SWE-bench Verified -- repository-level coding tasks}
\label{app:prompts-swebench}

On SWE-bench Verified, each method is expressed as a single
self-contained text prompt fed to the agent at task launch. The
prompts are inlined (rather than loaded via the agent's native skill
mechanism) so that every method's protocol is explicit and
reproducible. Every review-having prompt (AR and MARS) has the same
three-part shape:

\begin{itemize}[leftmargin=*,itemsep=0pt]
\item A \emph{shared SWE-bench wrapper} at the top
(\autoref{app:prompts-swebench-wrapper}) with repository state, issue
text slot, scope rules, and budget. Method-agnostic.
\item A \emph{method-specific body} -- AR's
\texttt{SKILL.md} of \autoref{app:prompts-swebench-ar-body} plus AR's
concrete SWE-bench instantiation of
\autoref{app:prompts-swebench-ar-workflow}, or the MARS
\texttt{SKILL.md} of \autoref{app:prompts-swebench-mars-body}.
\item A \emph{shared SWE-bench overrides} block at the bottom
(\autoref{app:prompts-swebench-overrides}) with the matched-model
directive and a final constraint reminder. Method-agnostic.
\end{itemize}

The Zero-shot prompt has only the shared wrapper -- no body, no
overrides. The AR row of \autoref{tab:swebench} uses the baseline AR
(two-verdict) protocol; the AR with text constraint upgrade of
\autoref{sec:prbench-v2} is not back-ported here. The MARS row uses
the same MARS protocol as in \autoref{sec:lcb-mar} and
\autoref{sec:prbench}: three independent reviewers, one
meta-reviewer, two outer rounds. The SWE-bench Verified-flavored
reviewer and meta-reviewer prompts are embedded inline inside the
\texttt{SKILL.md} body of \autoref{app:prompts-swebench-mars-body};
they have the same four-field output schema as the LCB and
SWE-PRBench MARS prompts, with the artifact slot replaced by
\texttt{git diff HEAD}.

\subsubsection{Zero-shot, MARS, and AR: top of every prompt (SWE-bench wrapper)}
\label{app:prompts-swebench-wrapper}
\label{app:prompts-swebench-ar-wrapper}

This block sits at the top of every method's task prompt. Slots
\texttt{\{repo\}}, \texttt{\{base\_commit\}}, and
\texttt{\{problem\_statement\}} are filled from the task instance.
The text is method-agnostic; everything method-specific lives in the
body that follows (\autoref{app:prompts-swebench-ar-body} for AR,
\autoref{app:prompts-swebench-mars-body} for MARS) and in the shared
overrides at the bottom (\autoref{app:prompts-swebench-overrides}).

\begin{footnotesize}
\begin{verbatim}
You are solving a real GitHub issue in an open-source Python
repository as part of a SWE-bench Verified evaluation.

## Repository state
- Repo: {repo}
- Commit: {base_commit}
- You are in a fresh checkout at this commit (CWD is the repo root).

## Issue to fix
{problem_statement}

## Your goal
Fix the issue by editing files in the repo. The runner captures
`git diff HEAD` after you exit -- that diff is what gets graded.

Common rules (apply to every method):
- Scope: fix exactly the issue described. Do NOT add features or
  refactor unrelated code.
- Do NOT edit test files (paths under tests/, test_*.py, *_test.py,
  testing/). The graders have their own hidden tests; touching the
  repo's tests contaminates evaluation.
- Do NOT commit. Leave changes uncommitted.
- Budget: ~30 minutes wall-clock total.

## CRITICAL -- do not load any preinstalled agent skill
The protocol-specific instructions that follow this block are
self-contained. Do NOT invoke any preinstalled agent skill. If you
do, this task is contaminated.
\end{verbatim}
\end{footnotesize}

\subsubsection{Zero-shot (SWE-bench Verified)}
\label{app:prompts-swebench-zeroshot}

The Zero-shot row of \autoref{tab:swebench} uses ONLY the shared
SWE-bench wrapper of \autoref{app:prompts-swebench-wrapper}. There
is no review-protocol body and no overrides block: the agent reads
the issue, edits the repo, and exits.

\subsubsection{MARS: \texttt{SKILL.md} (SWE-bench Verified)}
\label{app:prompts-swebench-mars-body}

The block below is MARS's \texttt{SKILL.md} as followed by the agent
on SWE-bench Verified. It sits between the shared wrapper
(\autoref{app:prompts-swebench-wrapper}) and the shared overrides
(\autoref{app:prompts-swebench-overrides}). The reviewer and
meta-reviewer prompts embedded inline below use the same four-field
output schema as their LCB
(\autoref{app:prompts-mars-reviewer-lcb},
\autoref{app:prompts-mars-meta-lcb}) and SWE-PRBench
(\autoref{app:prompts-mars-reviewer-prbench},
\autoref{app:prompts-mars-meta-prbench}) counterparts; the only
difference is the artifact framing (here, a unified diff captured by
\texttt{git diff HEAD}). The prompts are embedded inline so the agent
can follow the protocol without indirecting to another document.

\begin{footnotesize}
\begin{verbatim}
---
name: mars
description: Multi-Agent Review System (Wang et al. 2025). The author
generates a candidate solution. Three independent reviewers audit it.
A meta-reviewer aggregates the three reviews into a single decision
and revision suggestions. The cycle repeats up to K rounds.
---

# Multi-Agent Review System (MARS)

## Overview

MARS is a multi-agent review protocol with three role types: one
Author (you), three independent Reviewers, and one Meta-Reviewer.
The Author produces a candidate solution. The Reviewers each audit
the candidate independently and produce a structured review. The
Meta-Reviewer reads the candidate and the three reviews and issues
a single consolidated decision; on reject it also issues concrete
revision Suggestions. The Author revises and the cycle repeats.

Core principle: independent review without inter-reviewer
communication. Reviewers do NOT see each other's outputs. Diversity
comes from independent sampling on identical context, not from
debate.

Context rule: every reviewer and the meta-reviewer receives the same
full context -- the issue body, the current `git diff HEAD`, and
this protocol's prompt.

## When to Use

After the Author makes an initial fix. Before stopping.

## The MARS loop

### Step 1 -- Author produces a candidate fix

You are the Author. Read the repo, locate the bug, and make the
minimal fix via Edit.

### Step 2 -- Spawn 3 Reviewers IN PARALLEL

In one message, dispatch 3 Reviewer sub-agents via 3 parallel Task
tool calls. Each reviewer receives the issue body, the current
`git diff HEAD`, and the reviewer prompt below. Reviewers do NOT see
each other's outputs.

Reviewer prompt (verbatim):

  You are a reviewer in the MARS protocol. You evaluate a candidate
  solution INDEPENDENTLY of any other reviewer.

  Original ask:
  <ISSUE BODY>

  Candidate solution (unified diff):
  <GIT DIFF HEAD>

  Your task:
  1. Read the original ask and form an independent understanding of
     what a correct solution would do, BEFORE reading the candidate.
  2. Analyze the candidate against your understanding.
  3. Check: does the candidate satisfy the original ask? Are there
     edge cases missed? Does it introduce regressions? Is it scoped
     to the ask (no unrelated additions)?

  Output format (strict -- exactly these four labelled fields):

  Decision: [accept | reject]
  Confidence: [1-5]   (self-reported; 5 = certain)
  Justification: [evidence-based reasoning; cite file:line or code
                  snippets when applicable]
  Errors: [if reject: explicit list of errors in the candidate.
           If accept: leave blank.]

### Step 3 -- Spawn 1 Meta-Reviewer

After all three Reviewer sub-agents return, dispatch a single
Meta-Reviewer sub-agent. It receives the issue body, the current
`git diff HEAD`, and the three reviewer outputs.

Meta-Reviewer prompt (verbatim):

  You are the meta-reviewer in the MARS protocol. You received a
  candidate solution and three independent reviews of it. Your job
  is to issue a single consolidated decision plus, on reject,
  concrete revision suggestions for the author.

  Original ask:
  <ISSUE BODY>

  Candidate solution (unified diff):
  <GIT DIFF HEAD>

  Independent reviews:
  --- Reviewer 1 ---
  <REVIEW 1>

  --- Reviewer 2 ---
  <REVIEW 2>

  --- Reviewer 3 ---
  <REVIEW 3>

  Critical instructions:
  1. Do NOT count votes. Weigh the reasoning and evidence each
     reviewer provides.
  2. Confidence scores are signals, not deciders. A reviewer can be
     confidently wrong.
  3. If reviewers disagree, prefer the position grounded in concrete
     evidence about the candidate.

  Output format (strict -- exactly these four labelled fields):

  Decision: [accept | reject]
  Justification: [your reasoning, weighing the three reviews]
  Suggestions: [if reject: concrete, actionable revision
                instructions for the author. If accept: leave blank.]
  Recommended-answer: [a short summary of the correct solution as
                       you understand it]

### Step 4 -- Author revision

- If the Meta-Reviewer outputs Decision: accept, do a robustness
  pass: re-read your diff once more; if you spot a subtle bug, fix
  it. Otherwise leave the diff unchanged.
- If the Meta-Reviewer outputs Decision: reject, apply the
  Suggestions field via Edit. Do NOT restart from scratch -- revise
  the existing fix.

### Termination

Repeat Steps 2-4 for at most K = 2 rounds. After Round 2's revision
step, STOP. The runner captures `git diff HEAD`. Do not loop further.

## Why this structure

- Independent review without inter-reviewer communication: avoids
  debate-style overhead while preserving review diversity through
  independent sampling.
- Meta-Reviewer as decision authority: prevents simple vote counting;
  weighs evidence over numbers.
- Structured output schema: each role returns labelled fields, so
  the Author can act on them deterministically.

## Common Mistakes

- Spawning reviewers sequentially. Spawn them IN PARALLEL (one
  message with three Task tool calls). Sequential dispatch loses
  the independence guarantee if any state leaks across calls.
- Letting reviewers see each other. Each reviewer must receive only
  the original ask, the candidate, and the reviewer prompt -- never
  another reviewer's output.
- Skipping the Meta-Reviewer step. The Meta-Reviewer is the decision
  authority. Do NOT aggregate by hand from the three reviews --
  dispatch the Meta-Reviewer sub-agent.
- Vote counting inside the Meta-Reviewer. The Meta-Reviewer weighs
  evidence, not headcount.
- Exceeding K = 2 rounds. Stop after Round 2's revision step.
- Re-implementing from scratch on reject. Revise the existing fix
  using Suggestions; do not start over.
\end{verbatim}
\end{footnotesize}

\subsubsection{AR: \texttt{SKILL.md} (SWE-bench Verified)}
\label{app:prompts-swebench-ar-body}

The block below is AR's \texttt{SKILL.md}
and is embedded between the
\texttt{BEGIN VERBATIM SKILL CONTENT} and \texttt{END VERBATIM SKILL
CONTENT} markers shown above and below.

\begin{footnotesize}
\begin{verbatim}
---
name: adversarial-review
description: Use when changes are complete (code or plan docs) and the
agent is considering committing, proceeding with implementation, or
marking work as done. Also use when an implementation plan has been
drafted or modified before executing it.
---

# Adversarial Review

## Overview

Every change -- code or plan document -- must pass an adversarial
review loop before committing or proceeding. Three agent types must
agree: the main agent, a reviewer, and a critic of the reviewer. No
exceptions.

Core principle: Self-review is not review. A single reviewer is not
enough. Adversarial validation catches what consensus misses.

Context rule: Every sub-agent (reviewer AND critic) receives the same
full context -- diff, project docs (CLAUDE.md, architecture docs), and
related code. The critic is not a second-class citizen; it needs the
same information the reviewer had to properly evaluate the review.

## When to Use

Trigger on ANY of these:

- You finished implementing code changes and are about to commit
- You drafted or modified an implementation plan and are about to execute it
- You are about to mark a task or milestone as complete
- The user says "commit", "looks good", "ship it", or similar

This includes small or trivial changes, single-line fixes,
documentation-only changes, and plan documents before execution. No
size exemption.

Do NOT use for mid-implementation exploration (reading files, running
tests to understand behavior) or when the user explicitly says to skip
review for a specific change.

## The Adversarial Review Loop

CRITICAL STRUCTURE: The loop has two nested levels. The artifact
(code or plan) is ONLY edited by the main agent between outer
iterations -- never during the inner loop. Reviewer and critic
exchange text-only reviews; they never touch the artifact.

### Inner loop: converge on a consistent review

Given artifact version N (DO NOT edit the artifact during this loop):

1. Reviewer sub-agent evaluates version N with full context.
   Produces Review_1.
2. Critic sub-agent (fresh) evaluates Review_1 with the same full
   context the reviewer had.
   - If critic agrees with Review_k -> converged. Review_k is the
     consistent review.
   - If critic disagrees -> reviewer responds to the critique,
     producing Review_(k+1). Go to step 2 with the new review.
3. Continue bouncing until reviewer and critic agree on ONE review.

One inner round = reviewer response + critic response. Max 7 inner
rounds before escalating to the user.

The reviews exchanged here are text only. No edits to the code or
plan doc happen yet.

### Outer loop: iterate the artifact

After the inner loop converges on a consistent review:

Termination condition (strict): Terminate and commit/proceed if and
only if BOTH are true:

- The reviewer approved on their first pass -- Review_1 identified
  zero flaws
- The critic agreed on their first pass -- no back-and-forth in the
  inner loop

If terminated:
- If programmatic tests exist: run them, confirm they pass, then commit.
- If changes are UI-only / manually testable: ask the user to test on
  device, then commit.

Otherwise (the consistent review lists flaws, OR the inner loop
needed any back-and-forth to converge): the main agent edits the
artifact to address the flaws in the consistent review, producing
artifact version N+1. Start a fresh inner loop on version N+1.

### Why this structure

- Separation of review and editing prevents the main agent from
  prematurely editing based on one reviewer's opinion, before
  adversarial validation.
- First-pass termination guarantees quality: if the reviewer had to
  think twice, or the critic had to push back at all, the artifact
  isn't yet at a "both obviously approve it without discussion"
  state -- so it's not ready.
- Text-only inner loop keeps the artifact stable while reviewers
  figure out what (if anything) is actually wrong.

### Context for every sub-agent

Every sub-agent invocation (reviewer AND critic, every round, every
outer iteration) receives:

- The current artifact version (diff or plan document)
- Project docs (CLAUDE.md, architecture docs, related code)
- Any prior reviews in the current inner loop

Sub-agents are fresh -- they have no memory. Full context must be
passed every time.

## What the Reviewer Assesses

The reviewer is a senior software engineer evaluating: clarity,
consistency, feasibility, architecture, best practices, correctness,
security, testability.

## Two-Phase Workflow (Plans + Code)

When work involves a plan followed by implementation, the adversarial
review runs twice -- once on the plan, once on the code. You MUST NOT
collapse these into one review.

Phase 1 -- Plan review:
1. Write the plan document (or modify an existing one).
2. Run the full adversarial review loop on the plan.
3. Do NOT write any implementation code until reviewer + critic +
   main agent all agree on the plan.
4. If the review surfaces issues, revise the plan and re-run from
   step 2.

Phase 2 -- Code review:
1. Implement the agreed-upon plan.
2. Run the full adversarial review loop on the code changes.
3. Do NOT commit until reviewer + critic + main agent all agree on
   the implementation.
4. If the review surfaces issues, fix the code and re-run from step 2.

The sequence is: plan -> review plan -> agree -> implement -> review
code -> agree -> commit.

## Implementation Notes

- Use the Agent tool (not TaskCreate) to launch reviewer and critic
  sub-agents. Use subagent_type "general-purpose" or a code-reviewer
  agent type.
- Each sub-agent is fresh -- it has no memory of prior sub-agents.
  Always include full context in the prompt.
- If the original sub-agent can be continued via SendMessage, prefer
  that for the reviewer's follow-up responses to preserve
  conversation history.
- Launch reviewer and critic sequentially -- each depends on the
  previous output.

## Red Flags -- STOP and Review

You are about to violate this workflow if you think:

- "This change is too small to review"
- "I just wrote this, I know it's correct"
- "The user already approved it"
- "It's just a plan doc, not code"
- "Self-review of the diff is sufficient"
- "Running tests is enough without adversarial review"
- "The user is in a hurry"
- "This is a follow-up fix to reviewed code"
- "I'll implement first, then review the code"
- "The plan is straightforward, I can just implement it"
- "The reviewer already found the flaw, let me fix it while the
  critic reviews"
- "Small flaw, I'll just fix it now and the critic can catch up"
- "I agree with the reviewer, no need to wait for the critic's
  opinion"
- "The critic is going to agree, let me start editing"
- "Reviewer approved -- we're done" (without first-pass verification)
- "The critic disagreed but I agree with the reviewer, let me
  proceed"

All of these mean: start/continue the adversarial review loop. Do
NOT edit the artifact until the inner loop has converged on a
consistent review.

## Common Mistakes

- Editing the artifact during the inner loop. Reviewer and critic
  exchange reviews only. The artifact stays frozen until the inner
  loop converges. Premature edits defeat the whole point of
  adversarial validation.
- Terminating without first-pass verification. The only valid
  termination is "reviewer approved on Review_1 AND critic agreed
  immediately." Any back-and-forth means another outer iteration is
  required.
- Skipping the outer loop after edits. Once you edit the artifact,
  you MUST run a fresh inner loop. Prior approval of version N does
  not carry forward to version N+1.
- Skipping the critic. Sending only to one reviewer and committing
  on approval. The critic catches reviewer blind spots.
- Counting inner rounds wrong. One round is reviewer response +
  critic response. Reviewer alone is half a round.
- Shallow context. Sending just the diff without project docs.
  Reviewers and critics both need CLAUDE.md, architecture docs, and
  surrounding code -- every round.
- Committing before manual test confirmation. For UI-only changes,
  you MUST get explicit user confirmation from device testing.
\end{verbatim}
\end{footnotesize}

\subsubsection{AR: SWE-bench instantiation (concrete two-phase workflow)}
\label{app:prompts-swebench-ar-workflow}

This block sits between AR's \texttt{SKILL.md} body and the shared
overrides. It instantiates the abstract two-phase workflow from AR's
\texttt{SKILL.md} as a concrete SWE-bench routine, including the
plan-artifact location (APPROACH.md) and how the inner/outer loops
map to the Plan review and Code review phases. This block is
AR-specific; MARS has its own concrete workflow embedded in its
\texttt{SKILL.md} body (\autoref{app:prompts-swebench-mars-body}).

\begin{footnotesize}
\begin{verbatim}
## AR SWE-bench instantiation (applies AFTER the SKILL.md body above)

### Plan artifact location

Write your plan to `APPROACH.md` in the repo root (CWD).

`git diff HEAD` only shows changes to tracked files; new untracked
files don't appear. So APPROACH.md being untracked in the workdir is
invisible to the runner's diff capture and won't pollute the
prediction. The next task's setup runs `git clean -fdx` which removes
it.

One rule: do NOT `git add APPROACH.md`. Keep it untracked. If you
accidentally stage it, run `git rm --cached APPROACH.md` to untrack.

The plan should contain:
- Root cause -- what is actually wrong, with file:line references
- Proposed fix -- what change addresses the root cause (not just the
  symptom)
- Alternatives considered -- other approaches you ruled out and why
- Risk / edge cases -- what could break, how the fix handles edge
  cases
- Files to touch -- explicit list

### SWE-bench workflow (concrete form of the Two-Phase Workflow)

1. Explore the repo (Read, Grep, Glob, `git log`) -- read-only, no
   edits.
2. Write `APPROACH.md` in the workdir root (the plan). Keep it
   untracked.
3. Phase 1 -- Plan review (artifact = APPROACH.md):
   - Run the inner loop (max 5 inner rounds): R reviews APPROACH.md
     -> C evaluates R -> iterate to consistent review.
   - If first-pass clean, proceed to Phase 2.
   - Otherwise revise APPROACH.md, fresh inner loop. Max 2 outer
     iterations for Phase 1.
4. Implement the agreed plan (Edit). Make the minimal fix per
   APPROACH.md.
5. Phase 2 -- Code review (artifact = `git diff HEAD`):
   - Run the inner loop (max 5 inner rounds): R reviews the diff ->
     C evaluates R -> iterate to consistent review.
   - If first-pass clean, STOP.
   - Otherwise revise the diff, fresh inner loop. Max 2 outer
     iterations for Phase 2.
6. STOP.
\end{verbatim}
\end{footnotesize}

\subsubsection{Zero-shot, MARS, and AR: bottom of every prompt (SWE-bench overrides)}
\label{app:prompts-swebench-overrides}
\label{app:prompts-swebench-ar-overrides}

This block sits at the bottom of every review-having method's task
prompt (AR and MARS) after the method-specific body. It pins the
matched-model experiment and re-asserts the hard constraints from
the wrapper. Zero-shot does not include this block because Zero-shot
never spawns subagents.

\begin{footnotesize}
\begin{verbatim}
## Matched-model directive (applies to every Task tool call)

This is a matched-model experiment. The main agent and every
sub-agent (reviewer, critic, meta-reviewer, executor) MUST run on
Sonnet 4.5. When you spawn a sub-agent via the Task tool, pass:
  - subagent_type: "general-purpose"
  - model: "{model}"   (claude-sonnet-4-5-20250929 -- do NOT spawn
                        Opus or Haiku)

## Final constraints reminder

- Fix exactly the issue. Do NOT add features or refactor unrelated
  code.
- Do NOT edit test files (paths under tests/, test_*.py, *_test.py,
  testing/).
- Do NOT commit.
- Budget: ~30 minutes wall-clock total.
\end{verbatim}
\end{footnotesize}

\end{document}